\documentclass{elsart}

\makeatletter
\newif\if@restonecol
\makeatother

\usepackage{algorithmic}
\usepackage[linesnumbered,ruled,vlined]{algorithm2e}

\usepackage{graphicx}
\usepackage{amssymb}
\usepackage{amsmath}
\usepackage{enumerate}
\usepackage{multirow}
\usepackage{booktabs}
\usepackage[switch,pagewise]{lineno}
\usepackage[usenames]{color}
\usepackage{url}

\usepackage{rotating}

\usepackage{CJKutf8}
\usepackage[utf8]{inputenc}
\usepackage{bm}

\usepackage{threeparttable}
\begin{document}
\begin{CJK*}{UTF8}{gbsn}

\begin{frontmatter}

\title{Incorporating dictionaries into deep neural networks for the Chinese clinical named entity recognition}

%Authors and their institution
\author[ECUST]{Qi Wang},
\ead{dsx4602@163.com}
\author[ECUST]{Yuhang Xia},
\ead{153996626@qq.com}
\author[ECUST]{Yangming Zhou\corauthref{cor}},
\corauth[cor]{Corresponding author.}
\ead{ymzhou@ecust.edu.cn}
\author[ECUST]{Tong Ruan},
\ead{ruantong@ecust.edu.cn}
\author[ECUST]{Daqi Gao},
\ead{gaodaqi@ecust.edu.cn}
\author[SHDC]{Ping He}
\ead{2265137200@qq.com}
\address[ECUST]{Department of Computer Science and Engineering, East China University of Science and Technology, Shanghai, China}
\address[SHDC]{Shanghai Hospital Development Center, Shanghai, China}
\maketitle

\begin{abstract}

Clinical Named Entity Recognition (CNER) aims to identify and classify clinical terms such as diseases, symptoms, treatments, exams, and body parts in electronic health records, which is a fundamental and crucial task for clinical and translational research. In recent years, deep neural networks have achieved significant success in named entity recognition and many other Natural Language Processing (NLP) tasks. Most of these algorithms are trained end to end, and can automatically learn features from large scale labeled datasets. However, these data-driven methods typically lack the capability of processing rare or unseen entities. Previous statistical methods and feature engineering practice have demonstrated that human knowledge can provide valuable information for handling rare and unseen cases. In this paper, we address the problem by incorporating dictionaries into deep neural networks for the Chinese CNER task. Two different architectures that extend the Bi-directional Long Short-Term Memory (Bi-LSTM) neural network and five different feature representation schemes are proposed to handle the task. Computational results on the CCKS-2017 Task 2 benchmark dataset show that the proposed method achieves the highly competitive performance compared with the state-of-the-art deep learning methods.

\end{abstract}


\end{frontmatter}

% ------------------------------------------------------------------------------------------
\section{Introduction}
\label{Sec:Introduction}

Clinical Named Entity Recognition (CNER) is a critical task for extracting patient information from Electronic Health Records (EHRs) to support clinical and translational research. The main aim of CNER is to identify and classify clinical terms in EHRs, such as diseases, symptoms, treatments, exams, and body parts. There has been much work focused on extracting named entities from clinical texts, mostly because biomedical systems that rely on structured data are unable to access directly such healthcare information locked in the clinical texts, unless after CNER. However, building a CNER is not an easy task because of the richness of EHRs, and CNER in Chinese texts is more difficult compared to those in Romance languages due to the lack of word boundaries in Chinese and the complexity of Chinese composition forms \cite{Duan2011A}.

Recently, along with the development of deep learning methods, some neural network models \cite{Gridach2017Character,Habibi2017Deep} have also been successfully used for this task. Despite the great success achieved by deep learning methods, some issues still have not been well solved. One significant drawback is that such methods rarely take the integration of human knowledge into account. The deep neural networks usually employ an end-to-end approach and try to directly learn features from large scale labeled data. However, there also exists a huge number of entities that rarely or even do not occur in the training set. Several reasons lie behind this, including use of non-standard abbreviations or acronyms, multiple variations of same entities, etc. Thus, the data-driven deep learning methods usually cannot handle such cases well. While, dictionaries contain both commonly used entities and rare entities. If we can incorporate a dictionary into a deep neural network, rare and unseen clinical named entities can be better processed.

In this paper, we propose a novel method for the Chinese CNER. We extend Bi-directional Long Short-Term Memory and Conditional Random Field (Bi-LSTM-CRF) \cite{huang2015bidirectional} to model the CNER task as a character level sequence labeling problem. To integrate dictionaries, we design five different schemes to construct feature vectors for each Chinese character based on dictionaries and contexts. Also, two different architectures are introduced to integrate feature vectors with character embeddings to perform the task. Finally, our proposed approach was extensively evaluated based on a CCKS-2017\footnote{CCKS: China Conference on Knowledge Graph and Semantic Computing, 2017, and its website: http://www.ccks2017.com/} benchmark dataset.

The main contributions of this paper can be summarized as follows:
\begin{itemize}
\item To the best of our knowledge, it is the first time to incorporate dictionaries into deep neural networks for CNER tasks. We design two architectures and five feature representation schemes to integrate information extracted from dictionaries into deep neural networks.\\
\item We assess the performance of the proposed approaches on the CCKS-2017 Task 2 benchmark dataset. The computational results indicate that our proposed approaches perform remarkably well compared to state-of-the-art methods.
\end{itemize}

The rest of this paper is organized as follows. In the next section, we briefly review the related work of the clinical named entity recognition. Then we introduce the basic Bi-LSTM-CRF model in Section \ref{Sec:Bi-LSTM-CRF}. In Section \ref{Sec:Incorportating Dictionaries for Chinese CNER}, we present our proposed approaches. Section \ref{Sec:Computatioal Studies} is dedicated to some experimental studies. Finally, conclusions are provided in Section \ref{Sec:Conclusion}.

\section{Related work}
\label{Sec:Related Work}

Due to the practical significance, Clinical Named Entity Recognition (CNER) has attracted considerable research effort, and a great number of solution approaches have been proposed in the literature. Generally, all the existing approaches fall into four categories: rule-based approaches, dictionary-based approaches, statistical machine learning approaches, and recently, deep learning approaches are more investigated in CNER community.

Rule-based approaches rely on heuristics and handcrafted rules to identify entities. They were the dominant approaches in the early CNER systems \cite{Friedman1994A,Fukuda1998Toward} and some recent work~\cite{Zeng2006Extracting,Savova2010Mayo}. However, it is difficult to list all rules to model the structure of clinical named entities, and this kind of handcrafted approaches always leads to a relatively high system engineering cost.

Dictionary-based approaches rely on existing clinical vocabularies to identify entities \cite{rindflesch1999edgar,gaizauskas2000term,song2015developing}. They were widely used because of their simplicity and their performance. A dictionary-based CNER system can extract all the matched entities defined in a dictionary from a given clinical text. However, it cannot deal with entities which are not included in the dictionary, and usually causes low recalls.

Statistical machine learning approaches consider CNER as a sequence labeling problem where the goal is to find the best label sequence for a given input sentence \cite{Lei2014A,lei2014named}. Typical methods are Hidden Markov Models (HMMs) \cite{Zhou2002Named,song2015developing}, Maximum Entropy Markov Models (MEMMs) \cite{Mccallum2000Maximum,Finkel2004Exploiting}, Conditional Random Fields (CRFs) \cite{Mccallum2003Early,Settles2004Biomedical,Skeppstedt2014Automatic}, and Support Vector Machines (SVMs) \cite{Wu2006Extracting,Ju2011Named}. However, these statistical methods rely on pre-defined features, which makes their development costly. What's more, feature engineering, i.e. finding the best set of features which helps to discern entities of a specific type from others is more of an art than a science, incurring extensive trial-and-error experiments.

Recently, deep learning approaches \cite{Wu2015Named}, especially the methods based on Bidirectional Recurrent Neural Network (RNN) using CRF as the output interface (Bi-RNN-CRF) \cite{huang2015bidirectional}, achieve state-of-the-art performance in CNER tasks and outperform the traditional statistical models~\cite{Gridach2017Character,Habibi2017Deep,Zeng2017LSTM}. RNNs with gated recurrent cells, such as Long-Short Term Memory (LSTM) \cite{hochreiter1997long} and Gated Recurrent Units (GRU) \cite{Cho2014Learning}, are capable of capturing long dependencies and retrieving rich global information. The sequential CRF on top of the recurrent layers ensures that the optimal sequence of tags over the entire sentence is obtained.

\section{Bi-LSTM-CRF model}
\label{Sec:Bi-LSTM-CRF}

The Chinese clinical named entity recognition task is usually regarded as a sequence labeling task. Due to the ambiguity in the boundary of Chinese words, following our previous work \cite{Xia2017Clinical}, we label the sequence in the character level to avoid introducing noise caused by segmentation error. Thus, given a clinical sentence $X=<x_1,...,x_n>$, our goal is to label each character $x_i$ in the sentence $X$ with BIEOS (Begin, Inside, End, Outside, Single) tag scheme. An example of the tag sequence for ``腹平坦，未见腹壁静脉曲张。'' (The abdomen is flat and no varicose veins can be seen on the abdominal wall) can be found in Table \ref{table:labelingFormat}.

\begin{table}[!ht]
\begin{scriptsize}
\begin{center}
\caption{An illustrative example of the tag sequence and features.}
\label{table:labelingFormat}
\begin{threeparttable}
\begin{tabular}{c|c|c|c|c|c|c|c|c|c|c|c|c|c}
\toprule[0.75pt]
Character sequence  & \begin{CJK*}{UTF8}{gbsn}腹\end{CJK*}   & \begin{CJK*}{UTF8}{gbsn}平\end{CJK*} & \begin{CJK*}{UTF8}{gbsn}坦\end{CJK*} & \begin{CJK*}{UTF8}{gbsn}，\end{CJK*} & \begin{CJK*}{UTF8}{gbsn}未\end{CJK*} & \begin{CJK*}{UTF8}{gbsn}见\end{CJK*} & \begin{CJK*}{UTF8}{gbsn}腹\end{CJK*}   & \begin{CJK*}{UTF8}{gbsn}壁\end{CJK*}   & \begin{CJK*}{UTF8}{gbsn}静\end{CJK*}   & \begin{CJK*}{UTF8}{gbsn}脉\end{CJK*}   & \begin{CJK*}{UTF8}{gbsn}曲\end{CJK*}   & \begin{CJK*}{UTF8}{gbsn}张\end{CJK*}   & \begin{CJK*}{UTF8}{gbsn}。\end{CJK*} \\
\midrule[0.5pt]
tag sequence & S-b & O & O & O & O & O & B-b & E-b & B-s & I-s & I-s & E-s & O \\
\midrule[0.5pt]
PIET features & b & None & None & None & None & None & b & b & s & s & s & s & None \\
\midrule[0.5pt]
PDET features & S-b & None & None & None & None & None & B-b & E-b & B-s & I-s & I-s & E-s & None \\ \midrule[0.5pt]
entity type & body & \multicolumn{5}{c|}{} & \multicolumn{2}{c|}{body} & \multicolumn{4}{c|}{symptom} &  \\
\bottomrule[0.75pt]
\end{tabular}
\begin{tablenotes}
\tiny
\item[$\star$] The B-tag indicates the beginning of an entity. The I-tag indicates the inside of an entity. The E-tag indicates the end of an entity. The O-tag indicates the character is outside an entity. The S-tag indicates the character is merely a single-character entity. As for entity types, the b-tag indicates the entity is a body part, and the s-tag indicates the entity is a symptom.
\end{tablenotes}
\end{threeparttable}
\end{center}
\end{scriptsize}
\end{table}

\begin{figure}[!ht]
\begin{center}
\includegraphics[width=3.5in]{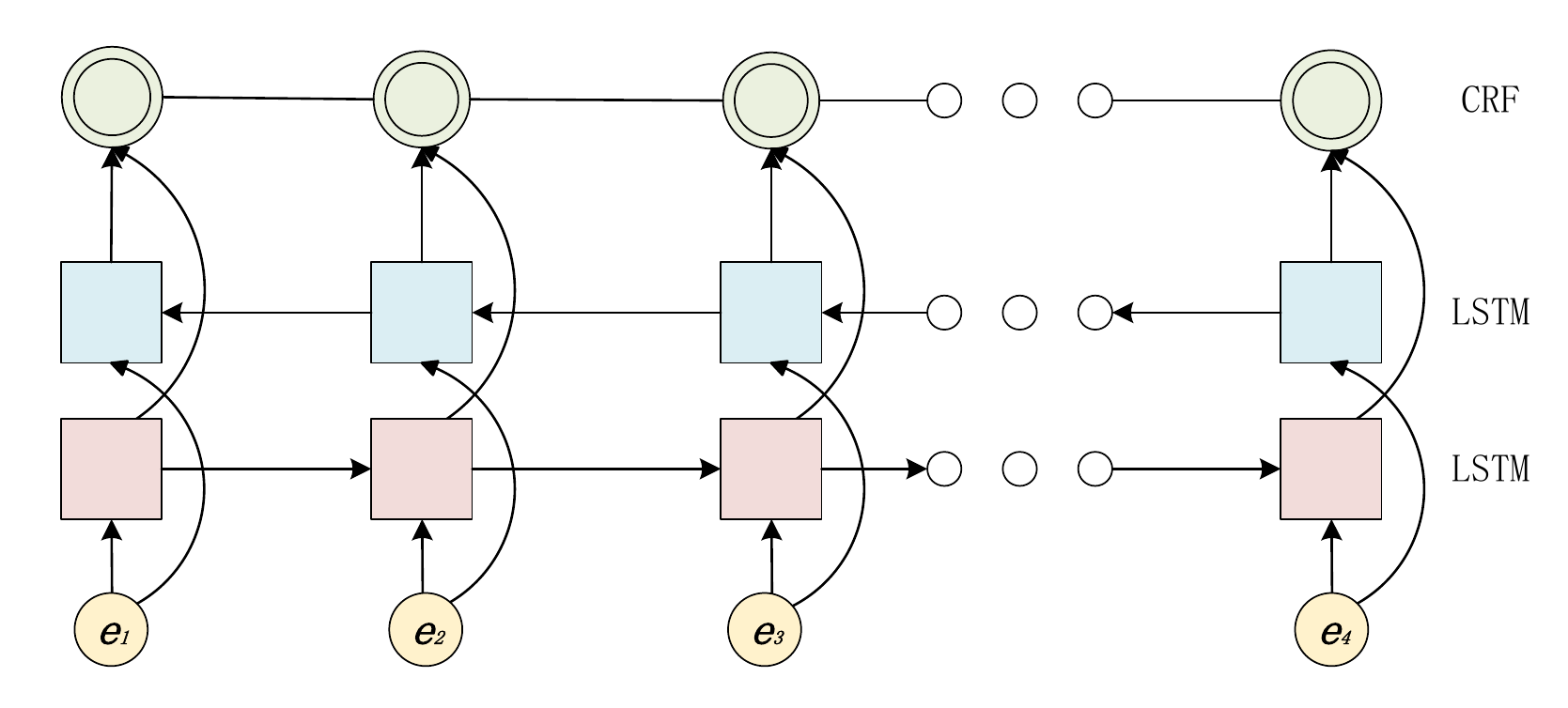}
\caption{Main architecture of the Bi-LSTM-CRF model.}
\label{Fig:Bi-LSTM-CRF}
\end{center}
\end{figure}

In this section, we will give a brief description of the general Bi-LSTM-CRF architecture for Chinese CNER. Bi-LSTM-CRF model is originally proposed by Huang \emph{et al.} \cite{huang2015bidirectional}, the main architecture of which is illustrated in Fig.~\ref{Fig:Bi-LSTM-CRF}. Unlike Huang \emph{et al.} \cite{huang2015bidirectional}, we employ character embeddings rather than word embeddings to deal with the ambiguity in the boundary of Chinese words.

\subsection{Embedding layer}

Given a clinical sentence $X=<x_1,...,x_n>$, which is a sequence of $n$ characters, the first step is to map discrete language symbols to distributed embedding vectors. Formally, we lookup embedding vector from embedding matrix for each character $x_i$ as $\bm{e}_i \in \mathbb{R}^{d_e}$, where $i \in \{1,2,...,n\}$ indicates $x_i$ is the $i$-th word in $X$, and ${d_e}$ is a hyper-parameter indicating the size of character embedding.

\subsection{Bi-LSTM layer}

The Long Short-Term Memory (LSTM) network \cite{hochreiter1997long} is a variant of the Recurrent Neural Network (RNN), which incorporates a gated memory-cell to capture long-range dependencies within the data and is able to avoid gradient vanishing/exploding problems caused by standard RNNs.

For each position $t$, LSTM computes $\bm{h_t}$ with input $\bm{e}_t$ and previous state $\bm{h}_{t-1}$, as:
\begin{align}
\bm{i}_t & = \sigma(\bm{W}_i\bm{e}_t+\bm{U}_i\bm{h}_{t-1}+\bm{b}_i)
\\
\bm{f}_t & = \sigma(\bm{W}_f\bm{e}_t+\bm{U}_f\bm{h}_{t-1}+\bm{b}_f)
\\
\tilde{\bm{c}}_t & = \tanh(\bm{W}_c\bm{e}_t+\bm{U}_c\bm{h}_{t-1}+\bm{b}_c)
\\
\bm{c}_t & = \bm{f}_t\odot \bm{c}_{t-1}+\bm{i}_t\odot \tilde{\bm{c}}_t
\\
\bm{o}_t & = \sigma(\bm{W}_o\bm{e}_t+\bm{U}_o\bm{h}_{t-1}+\bm{b}_o)
\\
\bm{h}_t & = \bm{o}_t\odot \tanh(\bm{c}_t)
\end{align}
where $\bm{h}$, $\bm{i}$, $\bm{f}$, $\bm{o} \in \mathbb{R}^{d_h}$ are $d_h$-dimensional hidden state (also called output vector), input gate, forget gate and output gate, respectively; $\bm{W}_i$, $\bm{W}_f$, $\bm{W}_c$, $\bm{W}_o \in \mathbb{R}^{4d_h \times d_e}$, $\bm{U}_i$, $\bm{U}_f$, $\bm{U}_c$, $\bm{U}_o \in \mathbb{R}^{4d_h\times d_h}$ and $\bm{b}_i$, $\bm{b}_f$, $\bm{b}_c$, $\bm{b}_o \in \mathbb{R}^{4d_h}$ are the parameters of the LSTM; $\sigma$ is the sigmoid function, and $\odot$ denotes element-wise production.

However, the hidden state $\bm{h}_i$ of LSTM only takes information from past, not considering future information. One solution is to utilize Bidirectional LSTM (Bi-LSTM)~\cite{Graves2005Framewise}, which incorporate information from both past and future. Formally, for any given sequence, the network computes both a left, $\overrightarrow{\bm{h}}_t$, and a right, $\overleftarrow{\bm{h}}_t$, representations of the sequence context at every input, $\bm{e}_t$. The final representation is created by concatenating them as:
\begin{align}
{\bm{h}_t}=\overrightarrow{\bm{h}}_t \oplus \overleftarrow{\bm{h}}_t
\end{align}

The Bi-LSTM along with the embedding layer is the main machinery responsible for learning a good feature representation of the data.

\subsection{CRF layer}

For the character-based Chinese CNER task, it is beneficial to consider the dependencies of adjacent tags. For example, a B (begin) tag should be followed by an I (middle) tag or E (end) tag, and an I tag cannot be followed by a B tag or S (single) tag. Therefore, instead of making tagging decisions using $\bm{h}_i$ independently, we employ a Conditional Random Field (CRF) \cite{Lafferty2001Conditional} to model the tag sequence jointly.

Generally, the CRF layer is represented by lines which connect consecutive output layers, and has a state transition matrix as parameters.  With such a layer, we can efficiently use past and future tags to predict the current tag, which is similar to the use of past and future input features via a Bi-LSTM network. We consider the matrix of scores $f_{\theta}([x]_{1}^{T})$ as the output of the Bi-LSTM network. The element $[f_{\theta}]_{i,t}$ of the matrix is the score output by the network with parameters $\theta$, for the sentence $[x]_{1}^{T}$ and for the $i$-th tag, at the $t$-th character.  We introduce a transition score $[A]_{i,j}$ to model the transition from $i$-th state to $j$-th for a pair of consecutive time steps.  Note that this transition matrix is position independent. We now denote the new parameters for our whole network as $\tilde{\theta}=\theta\cup\{[A]_{i,j}\forall i,j\}$. The score of $[x]_{1}^{T}$ along with a path of tags $[i]_{1}^{T}$ is then given by the sum of transition scores and Bi-LSTM network scores:
% \vspace{-2mm}
\begin{align}
	S([x]_{1}^{T},[i]_{1}^{T},\tilde{\theta})=\sum_{t=1}^T ([A]_{[i]_{t-1},[i]_{t}}+[f_{\theta}]_{[i]_{t},t})
\end{align}

The conditional probability $p([y]_{1}^{T}|[x]_{1}^{T},\tilde{\theta})$ is calculated with a softmax function:
\begin{align}
p([y]_{1}^{T}|[x]_{1}^{T},\tilde{\theta})=\frac{e^{S([x]_{1}^{T},[y]_{1}^{T},\tilde{\theta})}}{\sum_{j} e^{S([x]_{1}^{T},[j]_{1}^{T},\tilde{\theta})}}
\end{align}
where $[y]_{1}^{T}$ is the true tag sequence and $[j]_{1}^{T}$ is the set of all possible output tag sequences.

We use the maximum conditional likelihood estimation to train the model:
\begin{eqnarray}
\log \! p([y]_{1}^{T}|[x]_{1}^{T}\! ,\tilde{\theta}) \! = \! S([x]_{1}^{T} \! ,[y]_{1}^{T} \!,\tilde{\theta}) \! - \! \log \! \sum_{\forall [j]_{1}^{T}} \! e^{S([x]_{1}^{T} \! ,[j]_{1}^{T} \!,\tilde{\theta})}
\end{eqnarray}
Viterbi algorithm~\cite{Rabiner1990A}, which is a dynamic programming, can be used efficiently to compute $[A]_{i,j}$ and optimal tag sequences for inference.

\section{Incorporating dictionaries for Chinese CNER}
\label{Sec:Incorportating Dictionaries for Chinese CNER}

From the brief description given above, we can observe that the Bi-LSTM-CRF model can learn information from large-scale labeled data. However, it cannot process rare and unseen entities very well. Hence, in this work, inspired by the success of integrating dictionaries into the CRF models for CNER \cite{lin2007incorporating,Li2008Conditional}, we consider integrating dictionaries into the deep neural networks.

For a given sentence $X=<x_1,...,x_n>$, we first construct feature vector $\bm{d}_i$ for each character $x_i$ based on dictionary $D$ and the context. We propose five feature representation schemes for $\bm{d}_i$ to represent whether character segments that consist of character $x_i$ and its surroundings are clinical named entities or not. After that, we propose two architectures to integrate the feature vector $\bm{d}_i$ into the Bi-LSTM-CRF model. We will detail the feature vector construction and integration architectures in the following subsection.

\subsection{Feature vector construction}

In this section, we propose five different schemes to represent dictionary features. These schemes can be further classified into three categories: n-gram feature, Position-Independent Entity Type (PIET) feature and Position-Dependent Entity Type (PDET) feature.

\subsubsection{N-gram feature}

Given a sentence $X$ and an external dictionary $D$, we construct text segments based on the context of $x_i$ using the pre-defined n-gram feature templates. The templates used in our work are listed in Table~\ref{table:N-gramFeature}.

\begin{table}[!ht]
\begin{center}
\caption{N-gram feature templates for the $i$-th character, which are used to generate feature vector $\bm{d}_i$.}
\label{table:N-gramFeature}
\begin{tabular}{c|l}
\toprule[0.75pt]
Type   & template \\
\midrule[0.5pt]
2-gram & $x_{i-1}x_i$, $x_{i}x_{i+1}$\\
3-gram & $x_{i-2}x_{i-1}x_{i}$, $x_{i}x_{i+1}x_{i+2}$\\
4-gram & $x_{i-3}x_{i-2}x_{i-1}x_{i}$, $x_{i}x_{i+1}x_{i+2}x_{i+3}$\\
5-gram & $x_{i-4}x_{i-3}x_{i-2}x_{i-1}x_{i}$, $x_{i}x_{i+1}x_{i+2}x_{i+3}x_{i+4}$\\
\bottomrule[0.75pt]
\end{tabular}
\end{center}
\end{table}

\begin{figure}[!ht]
\begin{center}
\includegraphics[width=3.5in]{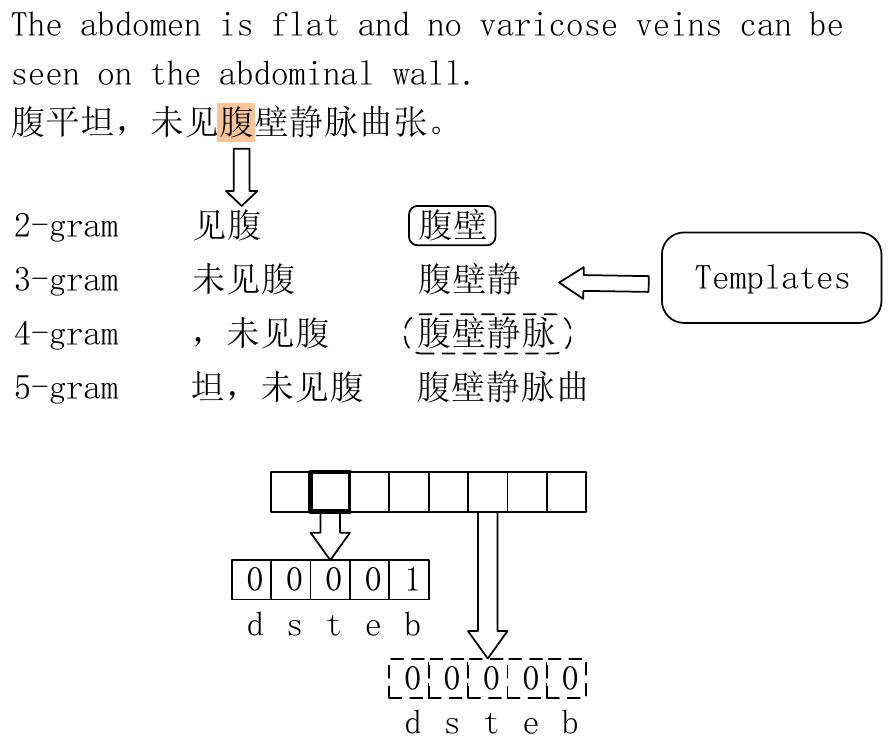}
\caption{Example of n-gram feature vector construction. The character with the red shadow is the character $x_i$. The character segment with solid rectangle is a body part in the dictionary $D$. Here d, s, t, e, b denote disease, symptom, treatment, exam, and body part, respectively.}
\label{fig:example}
\end{center}
\end{figure}

For a text segment that appears in an n-gram feature template, we can generate a binary value to indicate whether the text segment is a clinical named entity in $D$ or not. Here we use $t_{i,j,k}$ to represent the binary value of the output corresponding to the $k$-th entity type in $j$-th n-gram template for $x_i$. For a dictionary $D$ with five types of clinical named entities, we finally generate a 40-dimensional feature vector containing entity type and boundary information for $x_i$. Fig. \ref{fig:example} illustrates an example of n-gram feature vector construction.

\subsubsection{Position-Independent Entity Type (PIET) feature}

Given a sentence $X$ and an external dictionary $D$, we first use the classic Bi-Directional Maximum Matching (BDMM) algorithm \cite{gai2014} to segment $X$. The pseudo code of the BDMM algorithm is provided in Algorithm \ref{Alg:BDMM}. Then each character $x_i$ is labeled as the type of the entity which $x_i$ belongs to, as shown in the third line of Table \ref{table:labelingFormat}. The feature can be further represented in the format of one-hot encoding or feature embedding.

%\begin{algorithm}[!ht]
%\caption{The pseudo-code of the BDMM segmentation method}
%\label{Alg:BDMM}
%\begin{algorithmic}[1]
%\REQUIRE A clinical sentence $X$, a dictionary $D$ with different types of clinical named entities, and the maximum length $maxLen$ of the entities.
%\ENSURE The entity list $Y$ after segmentation\\
%\STATE $Y_1$= $\varnothing$;\\
%\STATE from the beginning to the end of $X$, each time we cut out a string $S$ whose length is $maxLen$\\
%\STATE let $S$ match with entities in $D$ in turn;\\
%\STATE if an entity matches $S$ then\\
%\STATE \quad split out $S$ from $X$\\
%\STATE \quad add $S$ with its type into $Y_1$\\
%\STATE \quad repeat the process above until finished\\
%\STATE else if we cannot find an entity matched in $D$
%\STATE \quad minus a character from the tail of $S$ and add the character back to $X$
%\STATE \quad continue to match $S$
%\STATE \quad if the length of $S$ is declined to 1 and still not matched then add $S$ with a specific type ``None'' that denotes no matched entities into $Y_1$
%\STATE end if
%\STATE $Y_2$= $\varnothing$;
%\STATE repeat the process above from the end to the beginning of $X$, and obtain the entity list $Y_2$
%\STATE end for\\
%\STATE set $Y$ as the one whose size is smaller between $Y_1$ and $Y_2$;
%\RETURN the entity list $Y$ after segmentation;
%\end{algorithmic}
%\end{algorithm}

\begin{algorithm}[ht]
\begin{small}
 \caption{The pseudo-code of the BDMM segmentation method}
 \label{Alg:BDMM}
\KwIn{A clinical sentence $X$, a dictionary with different types of clinical named entities $D$ and the maximum length of the entities in $D$ $maxLen$}
\KwOut{The entity list $Y$ after segmentation}
\Begin{
    $Y_1 \leftarrow \varnothing$;\\
    // \underline{Direction 1}: from the beginning to the end\\
    \While{$X$ is not empty}{
        cut a string $S$ of size $maxLen$ from $X$;\\
        make a match between $S$ and each entity in $D$;\\
        \If{there is an entity matches $S$}{
            split out $S$ from $X$;\\
            add $S$ with its type into $Y_1$;\\
            }
        \Else{
                remove a character from the tail of $S$;\\
                add the character back to $X$;\\
                make a match between $S$ and each entity in $D$;\\
                \If{a match is found}{
                    go to line 7;\\
                    }
                \If{$|S| = 1$}{
                    split out $S$ from $X$;\\
                    $Y_1 \leftarrow $``None'';\\
                }
                \Else
                {
                    repeat lines 11-17;
                }
            }
    }
    $Y_2 \leftarrow \varnothing$;\\
    // \underline{Direction 2}: from the end to the beginning\\
    build $Y_2$ in the same way of $Y_1$;\\
    \If{$|Y_1| < |Y_2|$}{
        $Y \leftarrow |Y_1|$;\\
        }
    \Else{
        $Y \leftarrow |Y_2|$;\\
        }
    }
\Return the entity list $Y$ after segmentation;
\end{small}
\end{algorithm}

\subsubsection{Position-Dependent Entity Type (PDET) feature}

PIET feature only considers the type of the entity which a character belongs to. Different from PIET feature, PDET feature also takes the position of a character in an entity into account: If the character is merely a single-character entity, we add a flag ``S'' before the PIET feature. Otherwise, for the first character of an entity, we add a flag ``B'' before the PIET feature; For the last character of an entity, we add a flag ``E'' before the PIET feature; For the middle character(s) of an entity, we add a flag ``I'' before the PIET feature. The example is shown in the fourth line of Table \ref{table:labelingFormat}. Similar to PIET feature, PDET feature can also be represented in the format of one-hot encoding or feature embedding.

To some degree, the feature vector, whatever n-gram feature or other entity type feature, can represent the candidate labels of a character based on the given dictionary. The values in the feature vector are dependent on the context and dictionary. They are not impacted by other sentences or statistical information. Hence, feature vectors can provide much information quite different from data-driven methods.

\subsection{Integration architecture}

Through the construction steps of dictionary feature vectors, given a sentence $X$, we obtain both character embedding $\bm{e}_i$ and feature vector $\bm{d}_i$ for each character $x_i$. As described above, the original Bi-LSTM-CRF model only takes $\bm{e}_i$ as inputs. Since dictionary features could provide valuable information for CNER, we try to integrate it with the original Bi-LSMT-CRF model. In this section, we will introduce two different architectures to integrate feature vectors with character embeddings.

\subsubsection{Model-I}

\begin{figure}[!ht]
\begin{center}
\includegraphics[width=3.5in]{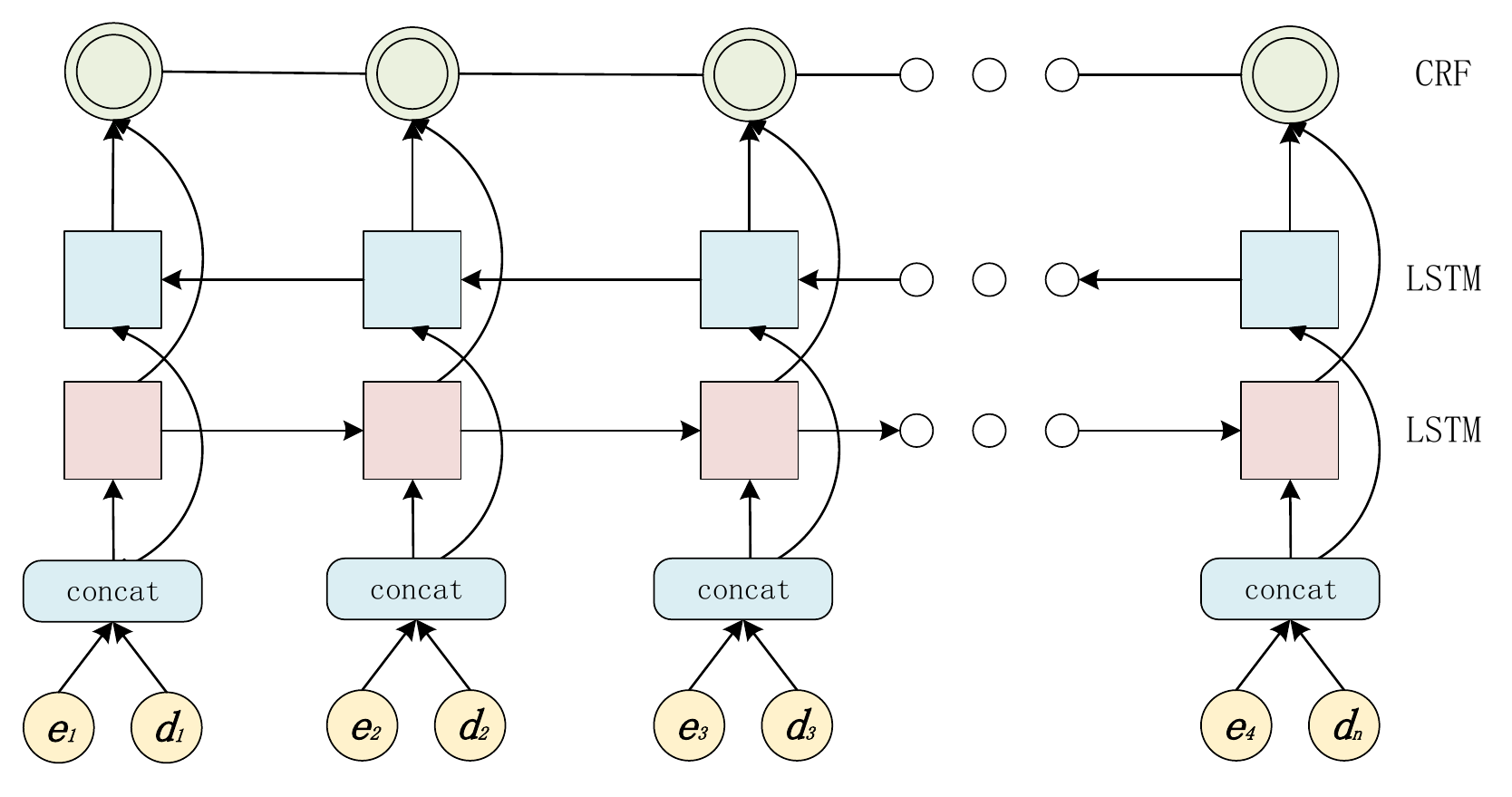}
\caption{Main architecture of Model-I.}
\label{Fig:model1}
\end{center}
\end{figure}

The general architecture of the proposed model is illustrated in Fig. \ref{Fig:model1}. The character embedding $\bm{e}_i$ and feature vector $\bm{d}_i$ are first concatenated and then fed into a Bi-LSTM layer:
\begin{equation}
\bm{m}_i = \bm{e}_i \oplus \bm{d}_i
\end{equation}
\begin{equation}
\bm{h}_i = \emph{\rm Bi-LSTM}(\overrightarrow{\bm{h}}_{i-1}, \overleftarrow{\bm{h}}_{i+1}, \bm{m}_i)
\end{equation}
where $\bm{e}_i$ denotes the embedding vector of $x_i$, and $\bm{d}_i$ represents the feature vector of $x_i$.

The other part of this model uses the same operation as the basic Bi-LSTM-CRF model.

\subsubsection{Model-II}

\begin{figure}[!ht]
\begin{center}
\includegraphics[width=3.5in]{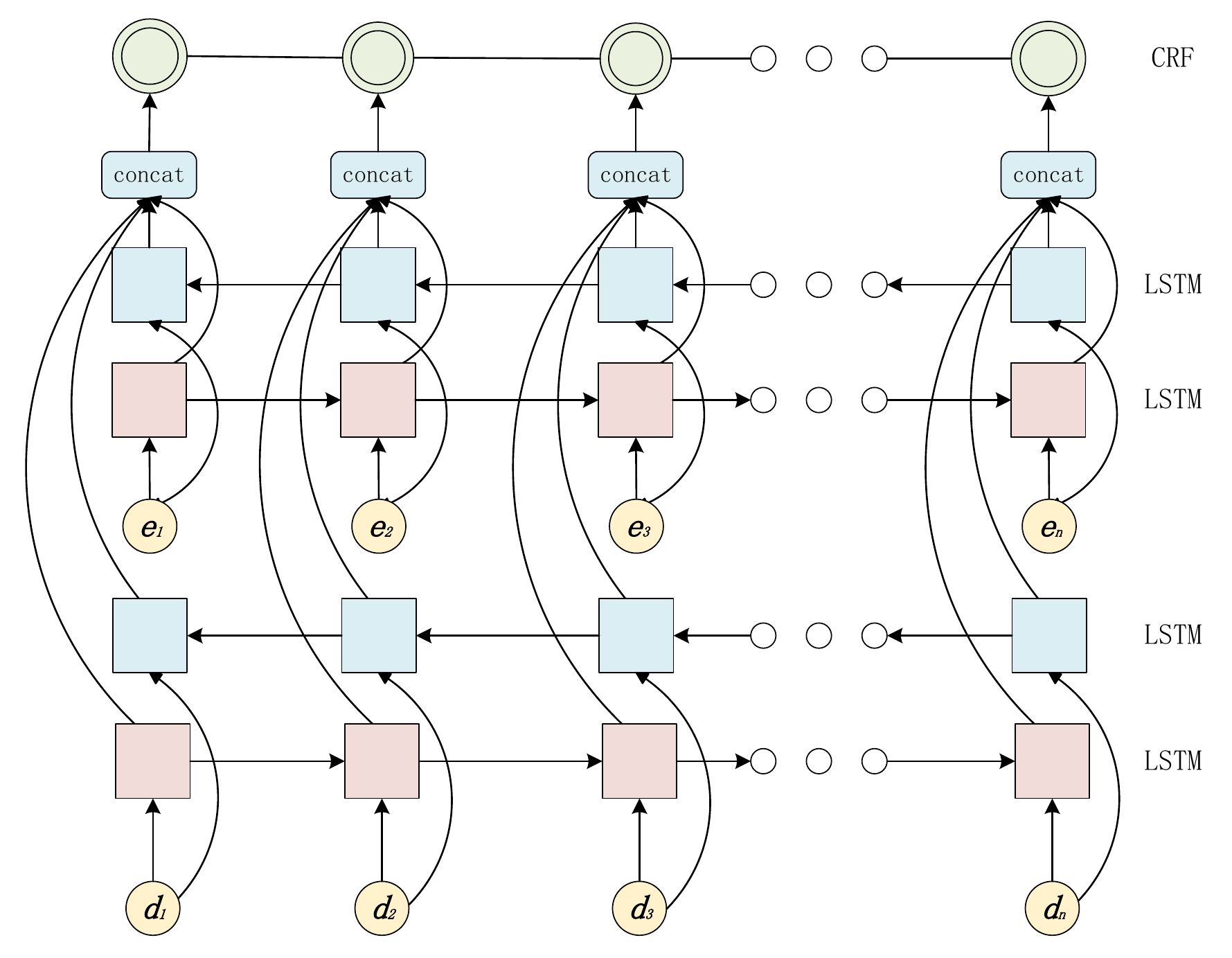}
\caption{Main architecture of Model-II.}
\label{Fig:model2}
\end{center}
\end{figure}

The general architecture of the proposed model is illustrated in Fig. \ref{Fig:model2}. The two parallel Bi-LSTMs can extract context information and potential entity information, respectively. For sentence $X$, the hidden states of the two parallel Bi-LSTMs at position $i$ can be defined as:
\begin{align}
\bm{h}_i^x & = \emph{\rm Bi-LSTM}(\overrightarrow{\bm{h}}_{i-1}^x, \overleftarrow{\bm{h}}_{i+1}^x, \bm{e}_i)
\\
\bm{h}_i^d & = \emph{\rm Bi-LSTM}(\overrightarrow{\bm{h}}_{i-1}^d, \overleftarrow{\bm{h}}_{i+1}^d, \bm{d}_i)
\end{align}
where $\bm{e}_i$ denotes the embedding vector of $x_i$, and $\bm{d}_i$ represents the feature vector of $x_i$. Note that in our formulation, the two parallel Bi-LSTMs are independent, without any shared parameters.

Finally, we concatenate the two hidden states of the parallel Bi-LSTMs as the inputs of the CRF layer:

\begin{eqnarray}
\bm{h}_i = \bm{h}_i^x \oplus \bm{h}_i^d
\end{eqnarray}

The other part of this model is the same as the basic Bi-LSTM-CRF model.

\section{Computational studies}
\label{Sec:Computatioal Studies}

The dictionary we exploit in the experiments is constructed according to the lists of charging items and drug information in Shanghai Shuguang Hospital as well as some medical literature such as \begin{CJK*}{UTF8}{gbsn}《人体解剖学名词（第二版）》\end{CJK*} (\emph{Chinese Terms in Human Anatomy [Second Edition]}).

We use the CCKS-2017 Task 2 benchmark dataset\footnote{It is publickly available at \url{http://www.ccks2017.com/en/index.php/sharedtask/}} to conduct our experiments. The dataset contains 1,596 annotated instances (10,024 sentences) with five types of clinical named entities, including diseases, symptoms, exams, treatments, and body parts. The annotated instances are already partitioned into 1,198 training instances (7,906 sentences) and 398 test instances (2,118 sentences). Each instance has one or several sentences. We further split these sentences into clauses by commas. The statistics of different types of entities are listed in Table \ref{table:dataset}.

We use the standard micro-average precision, recall, and F$_1$-Measure \cite{Liu2014A} to evaluate the methods in the following experiments.

\begin{table}[!ht]
\begin{center}
\caption{Statistics of different types of entities.}
\label{table:dataset}
\begin{tabular}{lrr}
\toprule[0.75pt]
Type& training set & test set \\
\midrule[0.5pt]
disease & 722 & 553\\
symptom & 7,831 & 2,311 \\
exam & 9,546 & 3,143 \\
treatment & 1,048 & 465 \\
body part & 10,719 & 3,021 \\
\hline
sum & 29,866 & 9,493 \\
\bottomrule[0.75pt]
\end{tabular}
\end{center}
\end{table}

\subsection{Experimental settings}

Parameter configuration may influence the performance of a deep neural network. The parameter configurations of the proposed approach are shown in Table \ref{Tab:Parameter Configuration}. Note that in order to avoid material impacts on the experimental results, the hidden unit number of each LSTM in Model-II is set the half of that in Model-I, so that the input length of CRF layer at each time step in Model-II is ensured the same as that in Model-I.

We initialize character embeddings and feature embeddings via word2vec \cite{mikolov2013efficient} on both the training data and the test data. Dropout technique \cite{Srivastava2014Dropout} is applied to the outputs of the Bi-LSTM layers in order to reduce over-fitting. The models are trained by Adam optimization algorithm \cite{kingma2014adam}, the parameters of which is the same as the default settings.

\begin{table}[!ht]
\begin{center}
\caption{Parameter configurations of the proposed approach.}
\label{Tab:Parameter Configuration}
\begin{tabular}{l|l}
\toprule[0.75pt]
Parameters          & value \\
\midrule[0.5pt]
character embedding size & $d_e$ = 128 \\
feature embedding size   & $d_d$ = 128 \\
number of LSTM hidden units in Model-I  & $d_h$ = 256 \\
number of LSTM hidden units in Model-II & $d_{h^x}$ = $d_{h^d}$ = 128 \\
dropout rate             & $p$ = 0.2 \\
batch size               & $b$ = 128 \\
\bottomrule[0.75pt]
\end{tabular}
\end{center}
\end{table}

\subsection{Comparative results of different combinations between feature representations and integration architectures}
\label{SubSec:Comparative Results of Different Combinations}

In this section, we compare different combinations between three feature representations and three integration architectures. The comparative results are listed in Table \ref{Tab:result}.

\begin{table}[!ht]
\begin{scriptsize}
\begin{center}
\caption{Comparative results of combinations between different feature representations and integration architectures.}
\label{Tab:result}
\begin{tabular}{c|cc|ccccccc}
\toprule[0.75pt]
\multicolumn{2}{c}{\multirow{2}{*}{}}  && \multicolumn{3}{c}{Model-I} && \multicolumn{3}{c}{Model-II} \\
\cline{4-6} \cline{8-10}
\multicolumn{2}{c}{} && Precision  & Recall       & F$_1$-Measure       && Precision        & Recall        & F$_1$-Measure   \\
\midrule[0.5pt]
\multicolumn{2}{c}{N-gram feature}  && 88.39    & 88.46   & 88.43   && 88.72    & 88.71    & 88.71   \\
\midrule[0.5pt]
\multirow{2}{*}{PIET feature}   & one-hot encoding   && 89.53    & 90.58   & 90.05   && 89.38    & 90.49    & 89.93   \\
& feature embedding && 90.11    & 90.01   & 90.56   && 90.00    & 90.60    & 90.30   \\
\midrule[0.5pt]
\multirow{2}{*}{PDET feature}   & one-hot encoding   && 90.51    & 91.04   & 90.77   && 90.22    & 90.64    & 90.43   \\
& feature embedding && \textbf{90.83}    & \textbf{91.64}   & \textbf{91.24}   && 90.36    & 91.35    & 90.85   \\
\bottomrule[0.75pt]
\end{tabular}
\end{center}
\end{scriptsize}
\end{table}

Table \ref{Tab:result} displays the computational results of each combination. First of all, Model-I with PDET features achieves the best performance among all the ten models, with $90.83\%$ in Precision, $91.64\%$ in Recall, and $91.24\%$ in F$_1$-Measure. Second, as for feature representations, n-gram features perform the worst compared to PIET features and PDET features, because n-gram features only consider the boundary information of potential entities, ignoring characters in the middle of entities. What's more, PDET features achieve the best results, because this type of features indicate not only the potential type but also the potential boundary of clinical named entities in the dictionary. Further more, as to PIET features and PDET features, feature embedding has better results than one-hot encoding. It is because the dense vector representation can bring more information than one-hot encoding. Third, except using n-gram features, Model-I performs better than Model-II in F$_1$-Measure, with an improvement of $0.28\%$ on average. It indicates that considering characters and their dictionary features together is better than considering them separately.

\subsection{Compared with two base models}

According to the comparative results of Section \ref{SubSec:Comparative Results of Different Combinations}, we know that the model combining Model-I and PDET features achieves the best performance. In this section, we compare the best model (Model-I with PDET features) with two base models, i.e. BDMM algorithm (see Algorithm \ref{Alg:BDMM}) and the basic Bi-LSTM-CRF model (see Section \ref{Sec:Bi-LSTM-CRF}). The comparative results are summarized in Table \ref{Tab:Compared With Base Models}.

\begin{table}[!ht]
\begin{center}
\caption{Comparative results between our best model and two base models.}
\label{Tab:Compared With Base Models}
\begin{tabular}{c|c|c|c}
\toprule[0.75pt]
     Method         & Precision     & Recall     & F$_1$-Measure\\
\midrule[0.5pt]
BDMM algorithm     & 70.29     & 84.44     & 76.72 \\
basic Bi-LSTM-CRF & 88.22     & 88.53     & 88.38 \\
our best model& \textbf{90.83} & \textbf{91.64} & \textbf{91.24} \\
\bottomrule[0.75pt]
\end{tabular}
\end{center}
\end{table}

From Table \ref{Tab:Compared With Base Models}, we clearly observe that our best model performs better than the two base models. It indicates the benefit of the incorporation between a dictionary and a Bi-LSTM-CRF model. BDMM algorithm with dictionaries performs the worst among the three models, and its Precision is far below its Recall. One reason is that applying BDMM algorithm directly may annotate clinical named entities with wrong boundaries. For example, ``双侧瞳孔'' (both pupils) is a body part in the clinical text, but it is not in the dictionary and the dictionary only has the body part ``瞳孔'' (pupil) in it, so the entity will be falsely recognized as ``瞳孔''. Another reason is type errors that the same entity can correspond to different entity types with different contexts. For example, ``维生素~C'' (vitamin C) is a drug name in the clause ``维生素C注射2g'' (inject with 2 grams of vitamin C), while it is also an exam index in the clause ``缺乏维生素~C'' (lack vitamin C). However, BDMM algorithm can only deal with the situation that an entity is correspond to one entity type.

\subsection{Compared with state-of-the-art deep models}

In this section, we compare our best model, i.e. Model-I with PDET features, with two state-of-the-art deep models in the literature. Li \emph{et al.} \cite{Li2017Recurrent} see the Chinese CNER task as a sequence labeling problem in word level. They exploit a Bi-LSTM-CRF model to solve the problem. To improve recognition, They also use health domain datasets to create richer, more specialized word embeddings, and utilize external health domain lexicons to help word segmentation.
Ouyang \emph{et al.} \cite{Ouyang2017Exploring} adopts bidirectional RNN-CRF architecture with concatenated n-gram character representation to recognize Chinese clinical named entities. They also incorporate word segmentation results, part-of-speech (POS) tagging and medical vocabulary as features into their model.
Hu \emph{et al.} \cite{Hu2017Compilation} develop a hybrid system based on rules, CRF and LSTM methods for the CNER task. They also utilize a self-training algorithm on extra unlabeled clinical texts to improve recognition performance. Note that except Li \emph{et al.} \cite{Li2017Recurrent}, the other systems all regard the CNER task as a character level sequence labeling problem.

The comparative results are shown in Table \ref{table:comparisonOthers}. From the table, we can see that our best model achieves the best results among all the models. Li \emph{et al.} \cite{Li2017Recurrent} get the worst performance because their word-level approach inevitably have wrong segmentation, leading to boundary errors when recognition, and the word set is much bigger than the character set, which means the corpus may be not big enough to learn word embeddings effectively. What's more, Hu \emph{et al.} \cite{Hu2017Compilation} utilize three separated models to handle the task, and finally get 91.02\% in F$_1$-Measure, which is the best one among the previous models, while we only exploit one model and achieve a 0.16\% improvement in F$_1$-Measure compared with Hu \emph{et al.} \cite{Hu2017Compilation}

\begin{table}[!ht]
\begin{center}
\caption{Comparative results between our best model with state-of-the-art deep models.}
\label{table:comparisonOthers}
\begin{threeparttable}
\begin{tabular}{c|c|c|c}
\toprule[0.75pt]
     Methods         & Precision     & Recall     & F$_1$-Measure\\
\midrule[0.5pt]
Li \emph{et al.} \cite{Li2017Recurrent}     & -     & -     & 87.95 \\
Ouyang \emph{et al.} \cite{Ouyang2017Exploring} & -     & -     & 88.85 \\
Hu \emph{et al.} \cite{Hu2017Compilation}     & \textbf{94.49} & 87.79 & 91.02 \\
Hu \emph{et al.} \cite{Hu2017Compilation}\tnote{$\star$}    & 92.99 & 89.25 & 91.08 \\
Our best model & 90.83 & \textbf{91.64} & \textbf{91.24} \\
\bottomrule[0.75pt]
\end{tabular}
\begin{tablenotes}
\tiny
\item[$\star$] The results are obtained by allowing the use of external resources for self-training.
\end{tablenotes}
\end{threeparttable}
\end{center}
\end{table}

\subsection{Parameter analysis}

In this section, we respectively investigate the impact of the dictionary size and the number of hidden units of LSTM on the performance of our approach. In following experiments, all experimental results are obtained by using our approach in the case of the Model-I and PDET features.

\subsubsection{The effect of the dictionary size}

\begin{figure}[!ht]
\begin{center}
\includegraphics[width=3.5in]{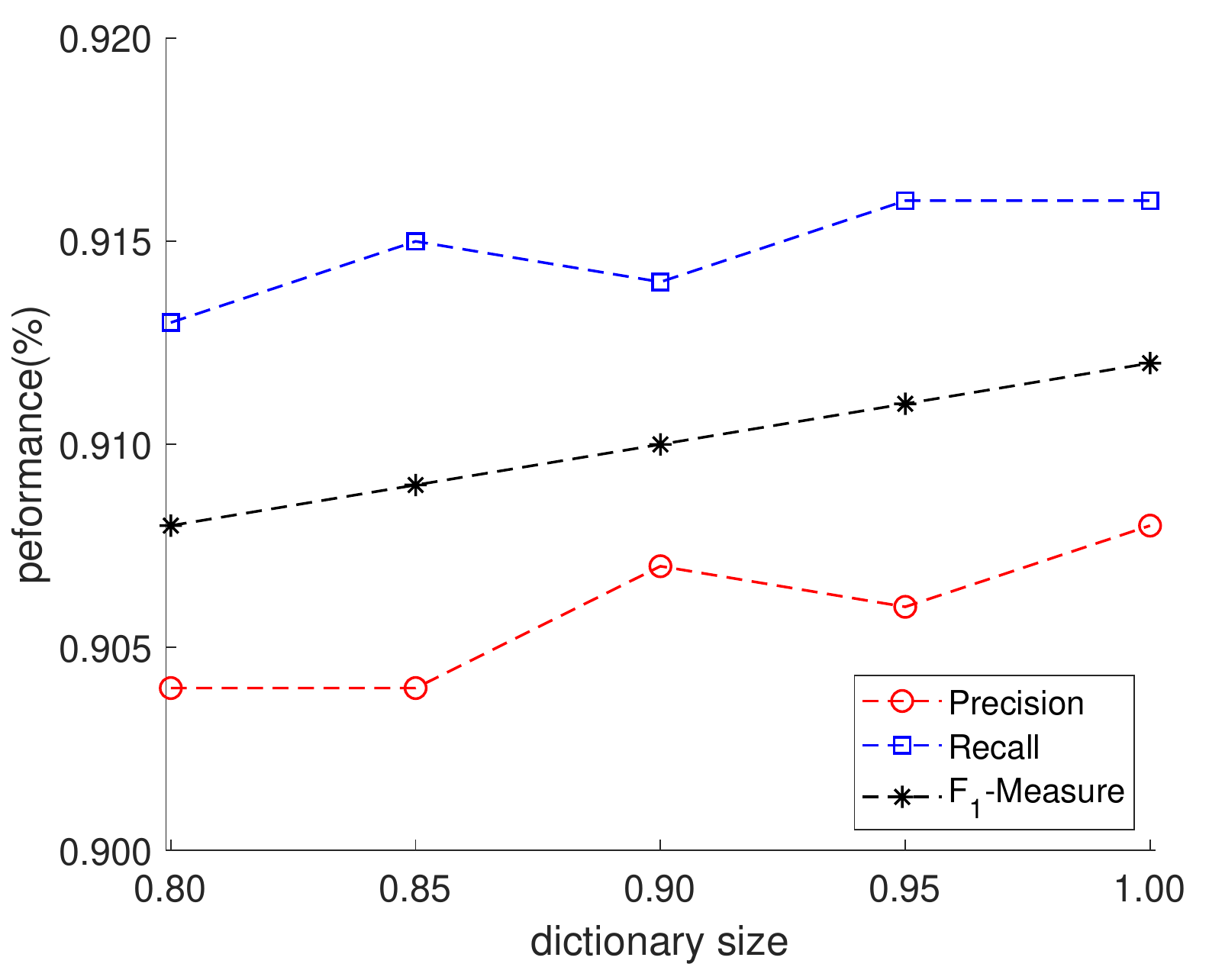}
\caption{The impact of the different dictionary size on the performance of our proposed approach in terms of Precision, Recall and F$_1$-Measure.}
\label{Fig:diffDicSize}
\end{center}
\end{figure}

We first investigate the effect of the dictionary size. We randomly select $80\%$, $85\%$, $90\%$, and $95\%$ of the entities from the original dictionary to construct new dictionaries with different sizes. Fig. \ref{Fig:diffDicSize} summarizes the computational results.

From Fig.~\ref{Fig:diffDicSize}, we can see that the performance of our proposed approach improves gradually as the dictionary size increases in all three different evaluation measures. In other words, if we build a dictionary containing more words, we can get better results.

\subsubsection{The effect of the hidden unit number}

\begin{figure}[!ht]
\begin{center}
\includegraphics[width=3.5in]{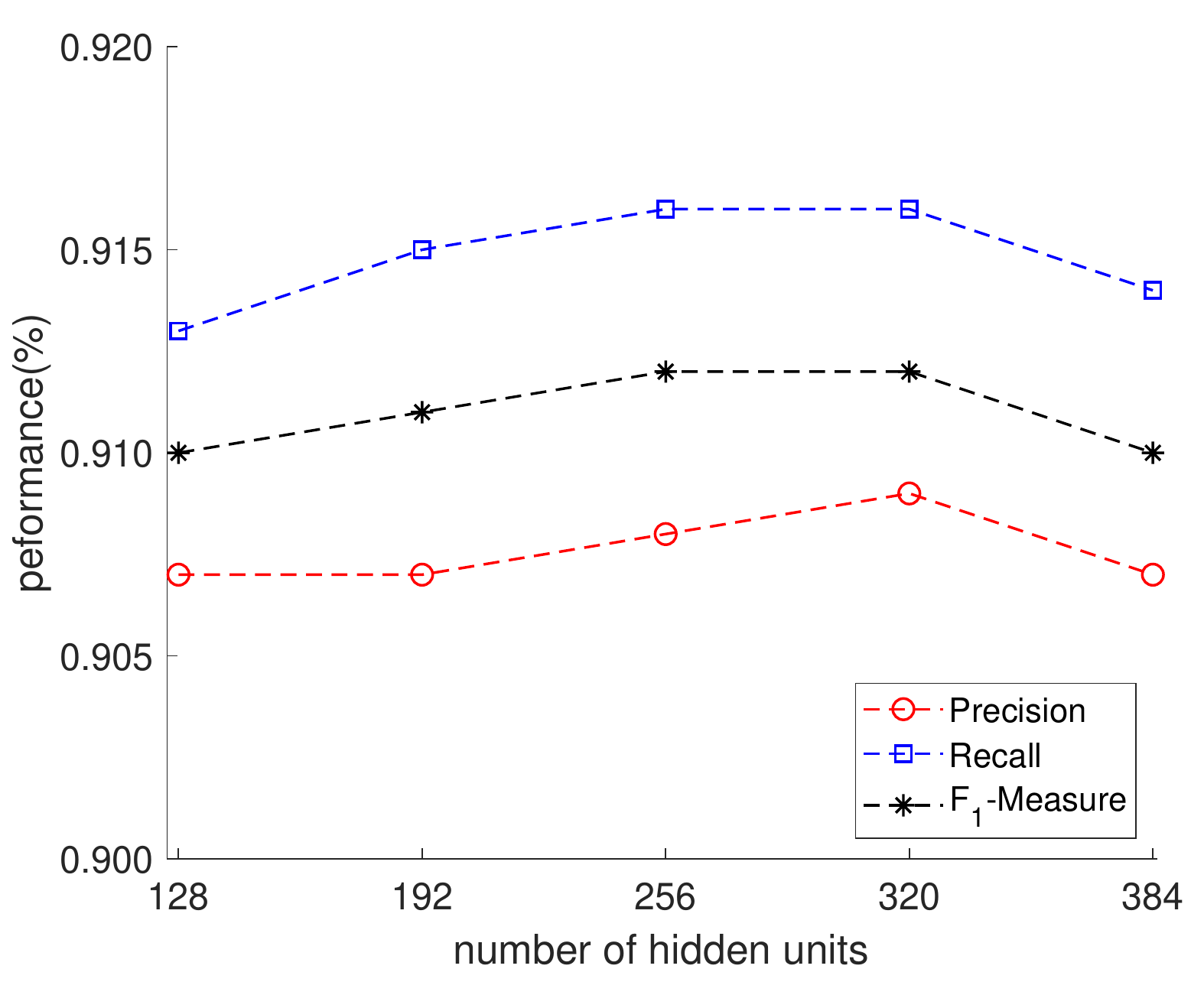}
\caption{The impact of different number of hidden units on the performance of our proposed approach in terms of Precision, Recall and F$_1$-Measure.}
\label{Fig:diffLSTMUnit}
\end{center}
\end{figure}

We further explore the influence of the hidden unit number $d_h$ of the LSTM, which takes the concatenation of character embeddings and PDET feature vectors as the inputs. In our experiment, $d_h$ is set to 128, 192, 256, 320, and 384, respectively. The results are shown in Fig. \ref{Fig:diffLSTMUnit}.

From Fig. \ref{Fig:diffLSTMUnit}, we can clearly observe that when the value of $d_h$ increases, the performance (in terms of Precision) of our proposed approach first grows a bit, then keeps stable, finally drop down. We can also obtain the same observations on the curves of Recall and $F_1$-Measure. These observations seem reasonable because a more complex model can obtain more powerful expression ability. However, the complexity of the model should match the amount of the training data. If the model is too complex, it will be subject to over-fitting and its generalization will be poor.

\section{Conclusion}
\label{Sec:Conclusion}

Since the previous methods do not take human knowledge into consideration when recognizing clinical named entities. In this paper, we propose effective approaches for the Chinese CNER by integrating dictionaries into neural network. Five different feature representation schemes are designed to extract information based on given dictionaries for clinical texts. Also, two different architectures are introduced to use the information extracted from the dictionaries. Due to dictionaries contain rare and unseen entities, the proposed approaches could process them better than previous methods.

Experimental results on the CCKS-2017 Task 2 benchmark dataset show that incorporated dictionaries could significantly enhance the performance of deep neural network for the Chinese CNER, and achieve highly competitive results compared with state-of-the-art deep neural network methods.

\section*{Acknowledgment}

This work is supported by the National Natural Science Foundation of China (No. 61772201), Scientific Research Program of Shanghai funded by Shanghai Science and Technology Committee (No. 16511101000), and Science and the Technology Innovation Project of Traditional Chinese Medicine funded by Shanghai Health and Family Planning Commission (No. ZYKC201601013).

\bibliography{bibfiles}
\end{CJK*}
\end{document}